\documentclass[journal=jctcce,manuscript=article]{achemso}

\usepackage{booktabs}
\usepackage[normalem]{ulem}
\useunder{\uline}{\ul}{}
\usepackage{url}
\usepackage{xr}
\usepackage{array}
\makeatletter
\newenvironment{datastatement}{%
\acs@section*{\datastatementname}%
}{}
\newcommand*\datastatementname{Data Availability Statement}
\makeatother

\author{Nicholas Casetti}
\affiliation{Department of Chemical Engineering, Massachusetts Institute of Technology, Cambridge, Massachusetts 02139, United States}
\author{Dylan Anstine} 
\affiliation{Department of Chemistry, Carnegie Mellon University, Pittsburgh, Pennsylvania 15213, United States}
\author{Olexandr Isayev} 
\affiliation{Department of Chemistry, Carnegie Mellon University, Pittsburgh, Pennsylvania 15213, United States}
\author{Connor W. Coley}
\affiliation{Department of Chemical Engineering, Massachusetts Institute of Technology, Cambridge, Massachusetts 02139, United States}
\altaffiliation{Department of Electrical Engineering and Computer Science, Massachusetts Institute of Technology, Cambridge, Massachusetts 02139, United States}
\email{ccoley@mit.edu}

\title{Anticipating the Selectivity of Intramolecular Cyclization Reaction Pathways with Neural Network Potentials}

\begin{document}

\newcommand{\crncaption}[1]{The chemical reaction network returned by REVAMP starting from \textbf{#1} where arrow color represents barrier height and node color represents intermediate energy relative to the reactant. Larger nodes correspond to species along the reaction pathway}

\newcommand{\doublecrncaption}[2]{The chemical reaction network returned by REVAMP starting from \textbf{#1} and \textbf{#2} respectively where arrow color represents barrier height and node color represents intermediate energy relative to the reactant. Larger nodes correspond to species along the reaction pathway}

\newcommand{\fullnetworkcaption}[1]{The chemical reaction network returned by REVAMP for case study #1 where arrow color represents barrier height. Solid arrows represent the experimentally observed reaction pathway.}

\newpage

\begin{abstract}

Reaction mechanism search tools have demonstrated the ability to provide insights into likely products and rate-limiting steps of reacting systems. However, reactions involving several concerted bond changes---as can be found in many key steps of natural product synthesis---can complicate the search process. To mitigate these complications, we present a mechanism search strategy particularly suited to help expedite exploration of an exemplary family of such complex reactions, cyclizations. We provide a cost-effective strategy for identifying relevant elementary reaction steps by combining graph-based enumeration schemes and machine learning techniques for intermediate filtering. Key to this approach is our use of a neural network potential (NNP), AIMNet2-rxn, for computational evaluation of each candidate reaction pathway. In this article, we evaluate the NNP's ability to estimate activation energies, demonstrate the correct anticipation of stereoselectivity, and recapitulate complex enabling steps in natural product synthesis.

\end{abstract}

\newpage

\section{Introduction} 

The computational modeling of reaction mechanisms provides unique characterization of various chemical processes. The understanding of mechanistic pathways and reaction networks yielded by these predictions have enabled, for example, deeper insight into fuel combustion \cite{sarathy_comprehensive_2011, keceli_automated_2019}, pathway elucidation for pyrolysis \cite{zhao_deep_2023, yang_generating_2024}, \emph{in silico} design of catalysts \cite{houk_computational_2008, raugei_toward_2015}, and---most related to this work---the assessment of feasibility in organic synthesis \cite{elkin_computational_2018, medina-franco_computational_2021, beker_prediction_2019, samha_exploring_2022, keto_data-efficient_2024, li_total_2025, rappoport_predicting_2019}.

Constructing mechanistic networks from a set of reactants is one method by which useful insights can be developed through computational analysis. These networks are generated by recursively identifying plausible mechanistic steps from some node in the chemical reaction network (starting at the reactants). Existing tools for automated mechanism search can be grouped into three categories based on how hypothetical elementary steps are proposed: physics-based, template-based, and graph-based. Physics-based approaches rely on continuous traversal of a potential energy surface (PES) to find transition states and intermediate products using electronic structure calculations such as density functional theory (DFT). This exploration can be driven by some artificial stimulus on the reactant molecule, such as an imposed force between specific atoms in AFIR \cite{maeda_communications_2010, maeda_artificial_2016}, geometric constraints along activating coordinates in IACTA \cite{lavigne_guided_2022}, or high temperature to encourage collisions in the Nanoreactor \cite{zhang_exploring_2024, wang_discovering_2014}. An advantage of such physics-based approaches is that all intermediates are guaranteed to have a continuous path from the reactant which creates a preference towards more relevant pathways. However, the iterative PES exploration necessary to provide an enumeration of possible intermediates as comprehensive as possible often requires evaluation of the forces of many (hundreds to thousands) of different atomistic configurations, which makes these methods computationally expensive when compared to alternatives. 

Template-based and graph-based approaches decouple the steps of \emph{proposing} elementary steps and \emph{evaluating} their likelihoods. Templates refer to  predefined reaction rules that define possible elementary steps from a given reactant. These elementary steps are then evaluated for kinetic/thermodynamic feasibility using any combination of heuristics, rate constant databases or predictions, or physics-based simulations (e.g., DFT via double-ended transition state search algorithms like nudged elastic band (NEB) \cite{jonsson_nudged_1998} and growing string method (GSM) \cite{zimmerman_growing_2013}). Template-based methods, among them NetGen \cite{broadbelt_computer_1994} and RMG \cite{gao_reaction_2016}, can hypothesize mechanisms more quickly than physics-based approaches but cannot predict elementary steps beyond the finite set of encoded reaction templates. Graph-based approaches overcome this limitation by enumerating modifications of a molecular graph to establish possible intermediates. These intermediates can then be evaluated with double-ended methods. Some examples of the graph-based approach are Z-Struct \cite{zimmerman_automated_2013}, ARD \cite{suleimanov_automated_2015}, and YARP \cite{zhao_simultaneously_2021}. The methods used by YARP, in particular, form the basis for the methods we employ in this work.

The primary drawback of graph-based methods is that combinatorial enumeration quickly becomes intractable as the number of atoms increases. Restrictions must be imposed to constrain the size of the candidate space, which may result in omission of important kinetic pathways. Furthermore, most graph enumeration schemes do not consider stereoisomerism, limiting  their applicability to the prediction of stereoselectivity. One reaction type where these drawbacks are particularly evident is cyclization reactions (e.g., cycloadditions, electrocyclizations). Cyclizations often involve multiple simultaneous bond breaking/forming events and often also operate with stereospecificity.
These reactions are crucial in the synthesis of many natural products. \cite{arns_cascading_2007, lobo_pericyclic_1997} Conveniently, they are also accurately modeled by gas phase DFT calculations \cite{houk_transition_1992}.

One common strategy for improving the computational efficiency of mechanism search approaches is to use a less expensive method than DFT to evaluate intermediates. For example, YARP uses GFN2-xTB \cite{bannwarth_gfn2-xtbaccurate_2019} to obtain transition state guesses \cite{zhao_simultaneously_2021}. IACTA also uses GFN2-xTB as the potential energy surface for coordinate scans to identify elementary steps \cite{lavigne_guided_2022}. Beyond semi-empirical methods, neural network potentials (NNPs) have been introduced to provide energy and force calculations with a drastically reduced computational cost compared to DFT. In these models, the relationship between chemical configuration and interatomic interactions is learned, primarily via neural networks, to reproduce energies and forces from reference quantum chemistry calculations \cite{kaser_neural_2023, anstine_machine_2023, behler_four_2021, zhang_roadmap_2025}. Though commonly trained on near-equilibrium structures, NNPs can be extended to cover reactive chemistry by specifically training on chemical configurations along minimum energy pathways that describe the transition from a reactant to a product  \cite{behler_first_2017}. Reactive NNPs have been successful in recapitulating DFT energies and geometries along minimum energy pathways for a variety of chemistries\cite{hu_neural_2023,manzhos_neural_2021,yoo_neural_2021, li_transition_2025, zhao_2_2023} and have been used with AFIR for mechanism search in Pd-catalyzed methane oxidation \cite{ichino_systematic_2024}. 

\begin{figure}[h]
    \centering
    \includegraphics[width=\textwidth]{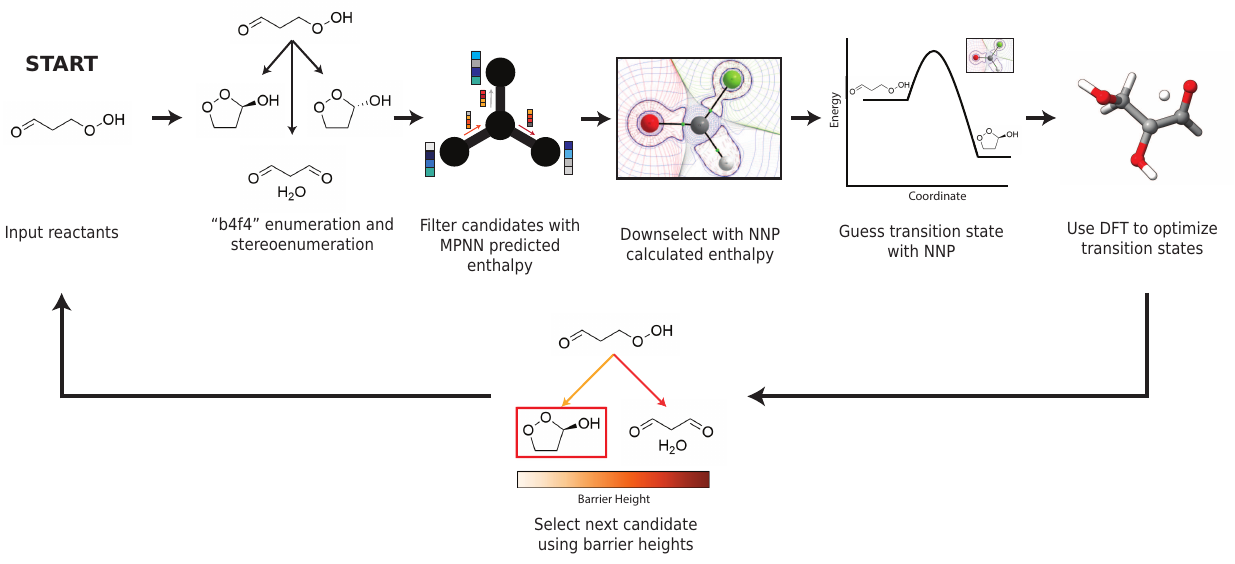}
    \caption{Overview of the methodology of REVAMP. REVAMP performs a b4f4 enumeration, including the use of stereoenumeration, and handles the larger resulting pool of intermediates by progressive filtering techniques using a combination of message passing neural networks (MPNNs) and a neural network potential (NNP).}
    \label{fig:overview}
\end{figure}

In this work, we illustrate the ability of an NNP-accelerated graph-based strategy for mechanism search, Reaction mechanism Exploration Via Automated Machine-learned Potential calculations (REVAMP; Figure~\ref{fig:overview}), to explore and accurately anticipate the selectivity of complex cyclization reaction pathways. Graph-based enumeration is paired with stereoenumeration to ensure a thorough consideration of potential elementary steps. To address the numerous intermediates that result from enumeration, unfavorable pathways are filtered with 2D message passing neural networks (MPNNs). A reactive NNP, a preliminary version of AIMNet2-rxn \cite{anstine_aimnet2-rxn_2025}, is then used to evaluate the remaining intermediates by removing excessively endothermic reactions before locating and optimizing transition states that are reevaluated by DFT. Key to this approach is AIMNet2-rxn which allows for sufficiently rapid evaluation of intermediates and calculation of transition states during a graph-based mechanistic search. In this study, the pre-trained NNP is restricted to neutral, closed-shell reactions consisting of C, H, N, and O atoms.

To demonstrate our method, we first benchmark AIMNet2-rxn in terms of its ability to recapitulate barrier heights and transition state geometries when compared to its reference method. We then demonstrate the utility of REVAMP with the representative use cases of exploring stereoselectivity and validating key steps of natural product syntheses. Ultimately, we demonstrate that pairing graph-based mechanism search with AIMNet2-rxn enables the successful exploration of complex reaction systems.

\section{Methods}

REVAMP requires the user to define the reactants of interest. The bond matrices of possible intermediates are obtained by enumerating graph rearrangements. This pool of candidates is downselected with increasingly stringent thermodynamic and kinetic filters to yield a set of elementary reaction steps with NNP-approximated transition states and barriers. The transition states of the lowest barrier reactions are then reoptimized using DFT. Carrying out this workflow recursively enables efficient construction of a reaction network. The specific NNP we employ in this study is a developmental version of AIMNet2-rxn trained on 2.3 million reference structures at the $\omega$b97M-V/def2-TZVPP level of theory which were generated as described in \citet{anstine_aimnet2-rxn_2025} Alternative NNPs trained on the recently released OMol25 dataset \cite{levine_open_2025} were demonstrated to be significantly more computationally expensive than AIMNet2-rxn and popular alternative GFN2-xTB (Section S7) making them less attractive for the methods employed in this study.

\subsection{Graph-based enumeration of hypothetical intermediate products}

The mechanism exploration in REVAMP begins with an enumeration of possible intermediate products given a starting chemical system (reactants).
We adopt a graph-based strategy which involves enumerating edits to a graph representation of the molecule. One common implementation uses the bond-electron matrix formalized by Ugi \cite{ugi_new_1979} and enumerates reaction matrices that represent the relocalization of valence electrons between the reactants the products \cite{suleimanov_automated_2015, zhao_simultaneously_2021, ramos-sanchez_automated_2023}. A different implementation, the one employed here, enumerates modifications to the bond connectivity matrix of the reactants \cite{zimmerman_automated_2013, ismail_automatic_2019}. It is possible to define all the possible modifications to the graph, but exhaustive enumeration is not tractable: in the simplified case of a binary bonding graph (where each of $N$ atoms can be connected to any other atom by a typeless edge), there are $2^{N(N-1)/2}$ bond rearrangements \cite{ismail_graph-driven_2022} which becomes intractable when $N > 10$. It is common to impose a ``b2f2'' constraint to limit the enumeration to elementary steps in which 2 bonds are broken and 2 bonds are formed; empirically, this is sufficient to capture many relevant chemical transformations and ensures a neutral system transforms into another neutral system without the introduction of radicals. However, it is not comprehensive enough to describe important classes of reactions including Diels-Alder, certain electrocyclizations, and Claisen rearrangements to name a few. Examples of some reactions described in the ``bnfn'' formalism can be seen in Figure S1. By default, REVAMP enumerates up to 4 bond rearrangements (``b2f2'', ``b3f3'', and ``b4f4'') to capture a broader scope of elementary steps.

\subsection{Rapid thermodynamic evaluation of hypothetical intermediate products}
 
Thermodynamic prescreening to filter out highly endothermic elementary steps is a common strategy for existing reaction network exploration approaches like YARP \cite{zhao_algorithmic_2022} as well as metabolic engineering applications \cite{kummel_putative_2006, henry_thermodynamics-based_2007, zamboni_annet_2008, kromer_constraining_2014}. 
We implement this strategy using reaction energy predictions from AIMNet2-rxn.
This evaluation involves generating a single intermediate conformer for each candidate and a reactant conformer with RDKit using the ETKDG method \cite{wang_improving_2020} and optimizing it with the NNP with the Fast Inertial Relaxation Engine (FIRE) \cite{bitzek_structural_2006} method in Pysisyphus \cite{steinmetzer_pysisyphus_2021}. The energy difference between each intermediate and the reactant is taken as the reaction energy.

One challenge with this evaluation is that single-point energy calculations---using an NNP or otherwise---require 3D conformers of reactant and product molecules. Conformer generation and optimization (energy minimization) can require on the order of 1 CPU second (Section S3). Our inclusion of b4f4 enumeration can increase the number of candidate intermediates by two orders of magnitude (Section S2) compared to b2f2, meaning that large molecules will yield very large numbers of reactions. We therefore benefit from the incorporation of a coarser thermodynamic feasibility filter at lower computational cost. In REVAMP, we introduce a learned filter that operates on the 2D graphs of molecules to prioritize speed for an initial triage at the risk of potentially missing thermodynamically relevant pathways. 

We train two directed message passing neural networks (MPNNs) \cite{yang_analyzing_2019, heid_chemprop_2024} for this purpose of thermodynamic feasibility assessment. One performs a binary classification of stability based on whether a molecule's adjacency matrix changes after geometry optimization. If the adjacency matrix changes during the geometry optimization, the molecule is labeled as unstable. The second uses a condensed graph of reaction (CGR) representation \cite{heid_machine_2022, hoonakker_representation_2011, varnek_substructural_2005} to predict the reaction energy, defined by the difference between the optimized energy of the reactant and product states. The training data for both models were generated by performing graph-based enumeration on 2100 neutral molecules containing only C, H, O, and N randomly sampled from reactants appearing in reactions from the United States Patent and Trademark Office (USPTO) \cite{lowe_extraction_2012}, randomly selecting 1000 of these enumerated intermediates for each, and using AIMNet2-rxn to evaluate their stability and reaction energy. These two models are used to evaluate candidate intermediates and remove molecules that are predicted to be unstable or highly endothermic (defaulting to 60 kcal/mol uphill). As the evaluation of an MPNN is far quicker than any pipeline involving conformer generation (ca. 1-5 CPU ms), this filter allows us to evaluate a much greater number of intermediates. Details regarding the performance of the MPNNs can be found in Section S5.

\subsection{Rapid kinetic evaluation of hypothetical intermediate products}

We treat thermodynamic feasibility as a necessary but not sufficient condition for an elementary step to be considered plausible. The next step in intermediate evaluation is estimating activation barriers of reactions that lead to the remaining intermediates. Rather than estimate activation energies from reactant and product graph structures alone \cite{grambow_deep_2020, ji_machine_2023, vadaddi_graph_2024, karwounopoulos_graph-based_2025}, we opt for a more robust evaluation of the already-narrowed space of possible reactions by efficient minimum energy pathway search with AIMNet2-rxn climbing image NEB calculations \cite{henkelman_climbing_2000}. 

Reactant-product conformers are generated using RDKit using the ETKDG method \cite{wang_improving_2020}. Specifically, reactant conformers are generated and optimized with the Universal Force Field (UFF) \cite{rappe_uff_1992}; then, the  reactant intramolecular geometry and force field parameters are updated to match the target product configuration. This product configuration is then reoptimized with UFF. The reactant-product pairs with the smallest RMSDs are then passed ahead to NEB calculations as this has shown to increase success rates of double-ended search methods when compared to random conformers \cite{zhao_conformational_2022}. It is important to note that this strategy does not guarantee the lowest energy TS will be obtained. Increasing the number of pairs (which defaults to 4) will increase the likelihood of finding the lowest energy TS conformer while also increasing computational expense. Image Dependent Pair Potential (IDPP) interpolation with 20 images is used to provide an initial guess for the NEB calculations  \cite{smidstrup_improved_2014}. Climbing image NEB is performed with Limited memory Broyden–Fletcher–Goldfarb–Shanno (LBFGS) optimization \cite{henkelman_climbing_2000, liu_limited_1989}. The highest energy NEB image is then used as the initial estimate for transition state optimization. Transition state optimization is performed by eigenvector following according to restricted step partitioned rational function optimization \cite{besalu_automatic_1998}. Converged transition states with a single imaginary frequency are then confirmed to correspond to reactant and product states through an intrinsic reaction coordinate (IRC) calculation \cite{fukui_formulation_1970, crehuet_reaction_2005}. All mechanistic modeling steps are performed using pysisyphus \cite{steinmetzer_pysisyphus_2021} with "gau\_vtight" convergence parameters. The default energetic threshold values used to filter intermediates for searches in this manuscript can be found in Section S6. These thresholds can be manually modified which gives the user control over the tradeoff of performing a more thorough search and reducing computational expense.

\subsection{Stereoenumeration of kinetically relevant pathways}

One pitfall associated with graph-based enumeration is that stereoisomers are implicitly ignored when operating on a 2D graph of the molecule unless stereochemistry is known \emph{a priori}, precluding analysis of stereoselectivity. We mitigate this issue by enumerating all possible stereoisomers of the intermediates that have been identified as potentially kinetically relevant in the previous step and evaluating their barriers. This is done by identifying all of the chiral centers and double bonds where stereoisomerism is possible (via RDKit) and exhaustively enumerating the possible combinations. Conformers for each stereoisomer are generated using ETKDG \cite{wang_improving_2020}, and instances where ETKDG fails to generate a conformer are labeled as physically unrealizable species. In practice, we find that evaluating all stereoisomers in this manner remains computationally tractable even for the complex case studies shown later.

\subsection{Refinement of activation barrier calculations with DFT}

For the final steps of REVAMP's evaluation pipeline, we reoptimize transition states, recalculate single point energies, and rerun the IRCs directly with DFT. Although AIMNet2-rxn can provide near-DFT accuracy \cite{anstine_aimnet2-rxn_2025}, REVAMP's broad enumeration scheme can lead to out-of-distribution structures far from AIMNet2-rxn's training set. AIMNet2-rxn predictions of energies for these out-of-distribution examples may exhibit different failure modes than DFT predictions, which tend to be better understood \cite{mardirossian_thirty_2017}. In the results reported below, we perform transition state optimizations and IRCs at $\omega$B97-X/6-31G and single-point calculations at $\omega$B97-X/6-311+G which has been demonstrated to yield reliable activation barriers for the reaction types explored in this manuscript by Houk and Maeda\cite{mita_prediction_2022}. All calculations were performed in Orca 4.2.1 \cite{neese_orca_2020} with default convergence parameters.

\subsection{Extension of single-step exploration strategy to mechanistic pathways}

To identify multi-step mechanisms, the single-step search is applied recursively. After identifying plausible elementary reaction products, one or more must be selected for further analysis, i.e., as the reactants in the subsequent step. REVAMP adopts from YARP a modified Dijkstra's algorithm where the barrier heights serve as edge weights to select the next reactant \cite{zhao_algorithmic_2022}. This exploration is run until there are no unexplored reaction pathways with a barrier below a certain energetic threshold---either predefined manually or defined relative to minimum energy pathways. The searches below use a threshold of 30 kcal/mol. We do this in lieu of calculating rate constants and using a flux-based criterion for termination \cite{han_--fly_2017, proppe_mechanism_2019, sumiya_rate_2020}. Future extensions to more complex reaction networks may benefit from these more robust search strategies as pruning based on flux, to the extent it can be estimated, can provide a more informative termination criteria than a fixed maximum barrier.

\section{Results}

\subsection{Validation of the accuracy of activation energies estimated by AIMNet2-rxn against its reference method}

\begin{figure}[h]
    \centering
    \includegraphics[width=\textwidth]{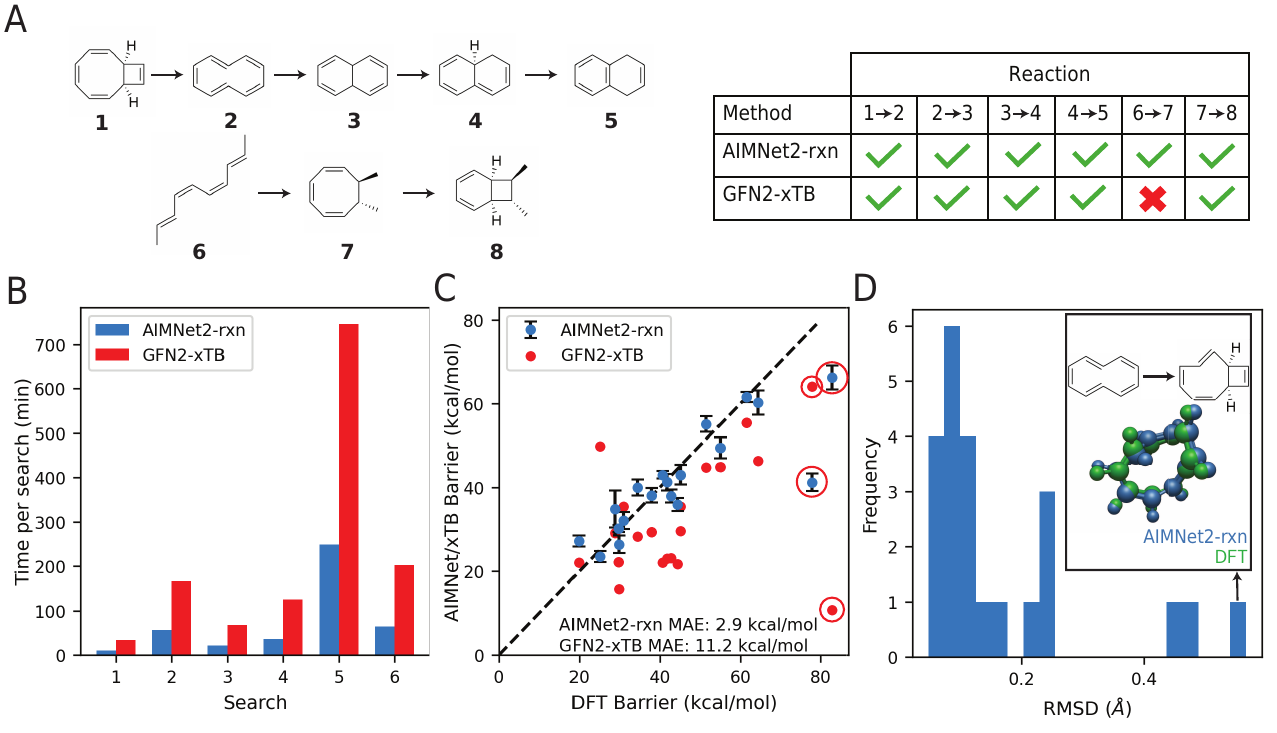}
    \caption{\textbf{A.} Two cyclization reaction mechanisms as explored by \citeauthor{mita_prediction_2022} \cite{mita_prediction_2022}. Using REVAMP with AIMNet2-rxn  finds transition states for all 6 mechanistic steps whereas REVAMP with GFN2-xTB misses an 8$\pi$ electrocyclization (\textbf{6 $\to$ 7}). \textbf{B.} Total wall time taken to perform kinetic feasibility calculations for AIMNet2-rxn and GFN2-xTB. \textbf{C.} Barrier heights as calculated by AIMNet2-rxn and GFN2-xTB compared to DFT. MAEs are calculated without circled outliers. Error bars are the sum of the standard deviation of the energies of the reactant and transition state between three AIMNet2-rxn model seeds.  \textbf{D.} Transition state RMSDs calculated between AIMNet2-rxn and DFT. The highest RMSD transition state is superimposed on the DFT transition state}
    \label{fig:benchmark}    
\end{figure}

We first explored the capability of AIMNet2-rxn to reproduce the results of its reference DFT method in our reaction space of interest. Two cascade cyclization reactions from \citeauthor{mita_prediction_2022} \cite{mita_prediction_2022} were selected as examples of complex rearrangements (Figure \ref{fig:benchmark}A). The first reaction transforms (1R,2Z,4Z,6Z,8S)-bicyclo[6.2.0]deca-2,4,6,9-tetraene (\textbf{1}) with a contrarotary 4$\pi$ decyclization followed by a disrotary 6$\pi$ cyclization. These are followed by two suprafacial [1,5]-hydrogen shifts to yield 1,4-dihydronaphthalene (\textbf{5}). The second reaction involves an 8$\pi$ then 6$\pi$ electrocyclization to form (1R,6S,7S,8S)-7,8-dimethylbicyclo-[4.2.0]octa-2,4-diene (\textbf{8}). We find that REVAMP is able to identify these two experimental reactive pathways using AIMNet2-rxn as described in Methods.

To benchmark AIMNet2-rxn against a common alternative, we replaced it with GFN2-xTB in the kinetic evaluation section of the workflow and repeated the mechanism searches. We found that GFN2-xTB was unable to localize a transition state for the 8$\pi$ electrocyclization (\textbf{6 $\to$ 7}) (Figure \ref{fig:benchmark}A). We also found that AIMNet2-rxn was able to perform the searches in less time (Figure \ref{fig:benchmark}B), though we note that both AIMNet2-rxn and GFN2-xTB were run within pysisyphus for these searches which likely reduced the efficiency of calculations. To evaluate the ability to recapitulate DFT barrier heights, we reoptimized the AIMNet2-rxn optimized reactants and transition states at $\omega$b97M-V/def2-TZVPP as this is the level of theory of AIMNet2-rxn's training data. An IRC was also performed to confirm that the transition state corresponded to the reaction found by the initial search. We also evaluate model uncertainty by recalculating the energy of the transition states and reactants with three AIMNet2-rxn models with different weight initializations. We find that AIMNet2-rxn is able to reproduce barrier heights better than GFN2-xTB (AIMNet2-rxn has a MAE of 2.9 kcal/mol, $R^2$ of 0.903 and GFN2-xTB has an MAE of 11.2 kcal/mol, $R^2$ of 0.406 excluding 2 outliers for both methods) (Figure \ref{fig:benchmark}C). We also find that in cases where AIMNet2-rxn is less accurate, there is often a higher level of model uncertainty (which is represented by the sum of the standard deviation of reactant and transition state energy between the three models). On top of the barrier height accuracy, AIMNet2-rxn is able to produce transition state geometries that closely resemble its DFT counterpart; the average RMSD between AIMNet2-rxn and DFT optimized transition state geometries was 0.17 $\r{A}$ with most geometries under 0.1 $\r{A}$ (Figure \ref{fig:benchmark}D). The transition state with the highest deviation from DFT is from a distorted conjugated system which yields a kinetically irrelevant product. These benchmarking results gave us confidence that the AIMNet2-rxn would faithfully reproduce DFT results in order to drive REVAMP searches.

\subsection{Accurate prediction of stereoselectivity in intramolecular Diels-Alder reactions}

\begin{figure}[h!p]
    \centering
    \includegraphics[width=\textwidth]{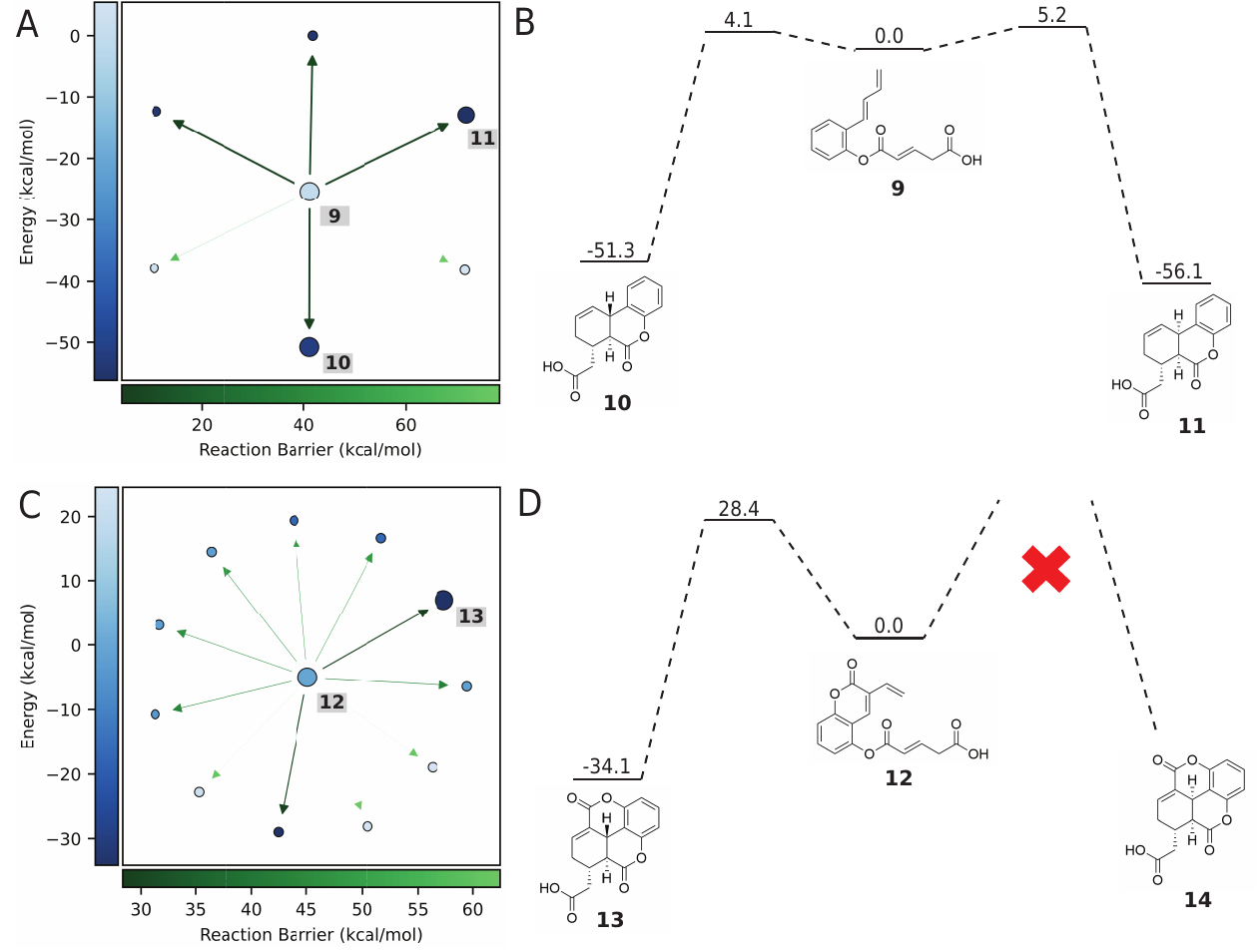}
    \caption{Successful prediction of intramolecular Diels-Alder stereoselectivity. \textbf{A,C.} \doublecrncaption{9}{12}. \textbf{B.} The cis and trans products of \textbf{9} are labeled with similar barriers matching experimental results of only weak preference for the trans stereoisomer. \textbf{D.} The extra steric hindrance associated with \textbf{12} leads REVAMP to prune the cis product on the basis of AIMNet2-rxn predictions, recapitulating the experimentally-observed selectivity of this reaction. The energies of each state are reported in kcal/mol normalized to the reactants.}
    \label{fig:da}
\end{figure}

One potential application of REVAMP is exploring the stereoselectivity of a reaction of interest. The stereoenumeration performed by REVAMP ensures that all possible stereoisomers of candidate intermediates are evaluated, allowing for an estimate of which is most favored. We demonstrate this capability by probing two different intramolecular Diels-Alder reactions (Figure \ref{fig:da}). Esters \textbf{9} and \textbf{12} are similar in structure but exhibit different enantioselectivities during intramolecular cyclization.

The first reaction of the unsubstituted fumarate ester (\textbf{9}) showed a 1.1 kcal/mol difference in barrier height  preferring the trans isomer (\textbf{10}), which aligns qualitatively with an experimental cis:trans ratio of 29:71 \cite{pearson_intramolecular_2006}. Contrarily, the C3-C12 tethered fumarate ester (\textbf{12}) was predicted to exhibit exclusively trans selectivity as AIMNet2-rxn filtered out the pathway leading to the cis product (\textbf{14}) for having an excessively high kinetic barrier. This once again aligns with experimental results as only the trans isomer was able to be detected by crystal X-ray analysis \cite{pearson_controlling_2008}.

\subsection{Retrospective analysis of key steps in natural product synthesis planning}

In natural product synthesis, there is often an enabling reaction that enables the construction of the molecular scaffold along the synthesis route; for a subset of products, this may be a cyclization that simultaneously sets multiple stereocenters.\cite{heravi_recent_2015} 
In recent years, detailed computational analysis of proposed synthetic steps have been used to explain selectivity and guide synthetic efforts \cite{elkin_computational_2018, medina-franco_computational_2021, beker_prediction_2019, samha_exploring_2022, keto_data-efficient_2024, li_total_2025, rappoport_predicting_2019}, thus we see an opportunity to apply REVAMP to the assessment of feasibility of these enabling reactions during natural product route planning as a way to derisk experiments and reveal potential unwanted byproducts.

\subsubsection{Synthesis of a key salvinorin intermediate}

\begin{figure}[h]
    \centering
    \includegraphics[width=\textwidth]{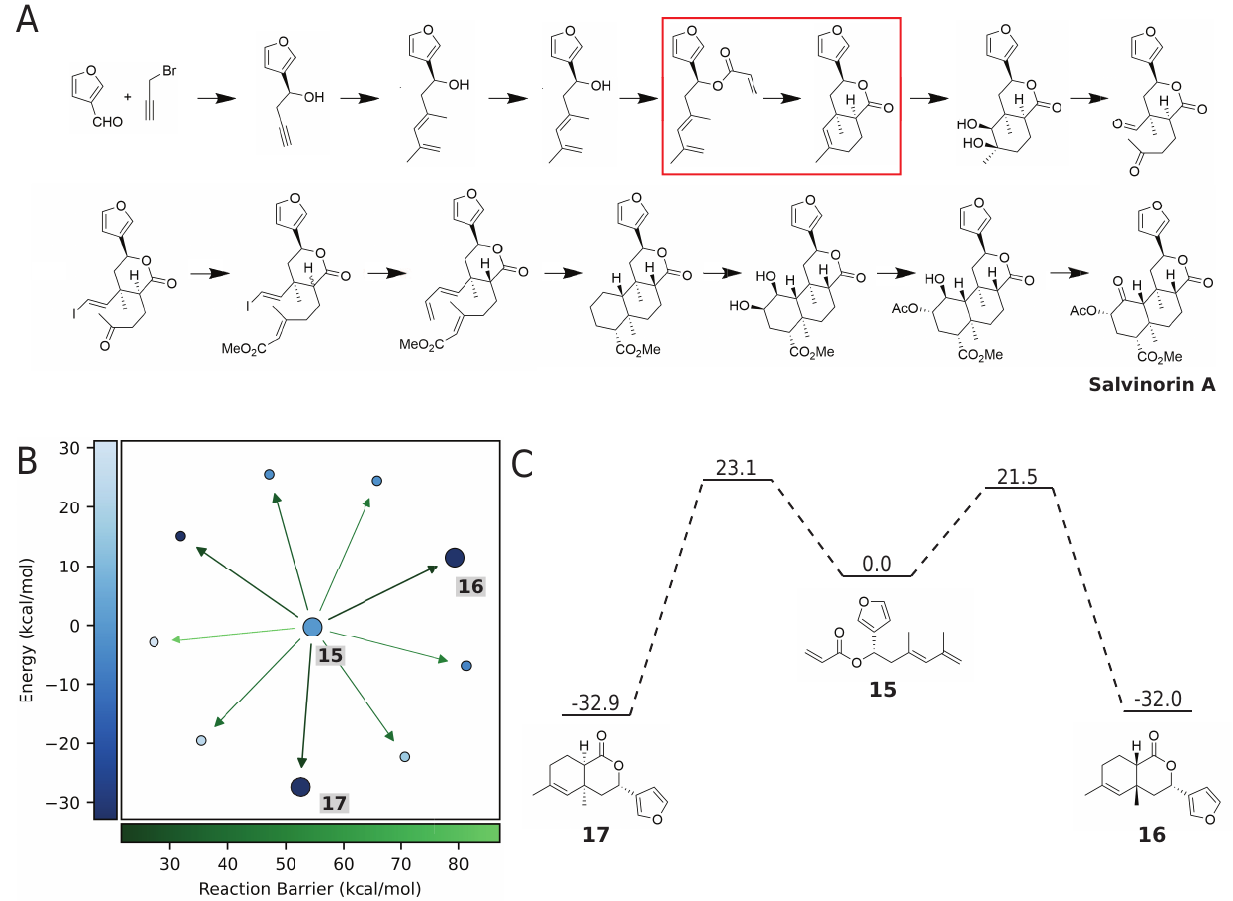}
    \caption{Analysis of the synthesis of salvinorin A. \textbf{A.} The full synthesis route from \citeauthor{zimdars_protectinggroupfree_2021} \cite{zimdars_protectinggroupfree_2021}; we apply REVAMP to evaluate the key step boxed in red through a mechanism search. \textbf{B.} \crncaption{15}. \textbf{C.} The lowest barrier step found by REVAMP is consistent with the synthesis route from \citeauthor{zimdars_protectinggroupfree_2021}. The energies of each state are reported in kcal/mol normalized to the reactants.}
    \label{fig:salvinorin}
\end{figure}

As our validations thus far have demonstrated that REVAMP is particularly compatible with the evaluation of Diels-Alder cycloadditions, the first synthesis planning we explored was a key step of Zimdars' synthesis of salvinorin A \cite{zimdars_protectinggroupfree_2021} (Figure \ref{fig:salvinorin}A). In this synthesis, a single concerted step enables the generation of two cycles and two stereocenters. The preferred pathway found in our analysis is the intramolecular Diels-Alder reaction of (S,E)-1-(furan-3-yl)-3,5-dimethylhexa-3,5-dien-1-yl acrylate (\textbf{15}) to a lactone (\textbf{16}) (Figure \ref{fig:salvinorin}C) which matches the experimental results from \citeauthor{zimdars_protectinggroupfree_2021} \cite{zimdars_protectinggroupfree_2021}.

REVAMP identifies two other diastereomers (Figure S11) of the same lactone with one of the diastereomers (\textbf{17}) having a barrier only 1.6 kcal/mol greater than the product (\textbf{16}) which would indicate this diastereomer would appear as a minor product. However, this step is experimentally reported to be completely selective for \textbf{16} indicating minor disagreement between experiment and our calculated results. Despite this, the major product for this reaction is still identified as the preferred pathway in our analysis.

\subsubsection{Synthesis of endiandric acid C}

Key steps that span multiple elementary steps, as in cascade reactions, may be more challenging to predict \emph{a priori} through intuition alone. We performed a mechanistic search starting from the complex polyene \textbf{18} from the last step in the Nicolaou's synthesis of endiandric acid C, which involves three sequential cyclization reactions \cite{nicolaou_endiandric_1982}. These elementary steps comprise a b4f4 cyclization as well as a stereodetermining step that yields a bridged ring.

\begin{figure}[hp]
    \centering
    \includegraphics[width=\textwidth]{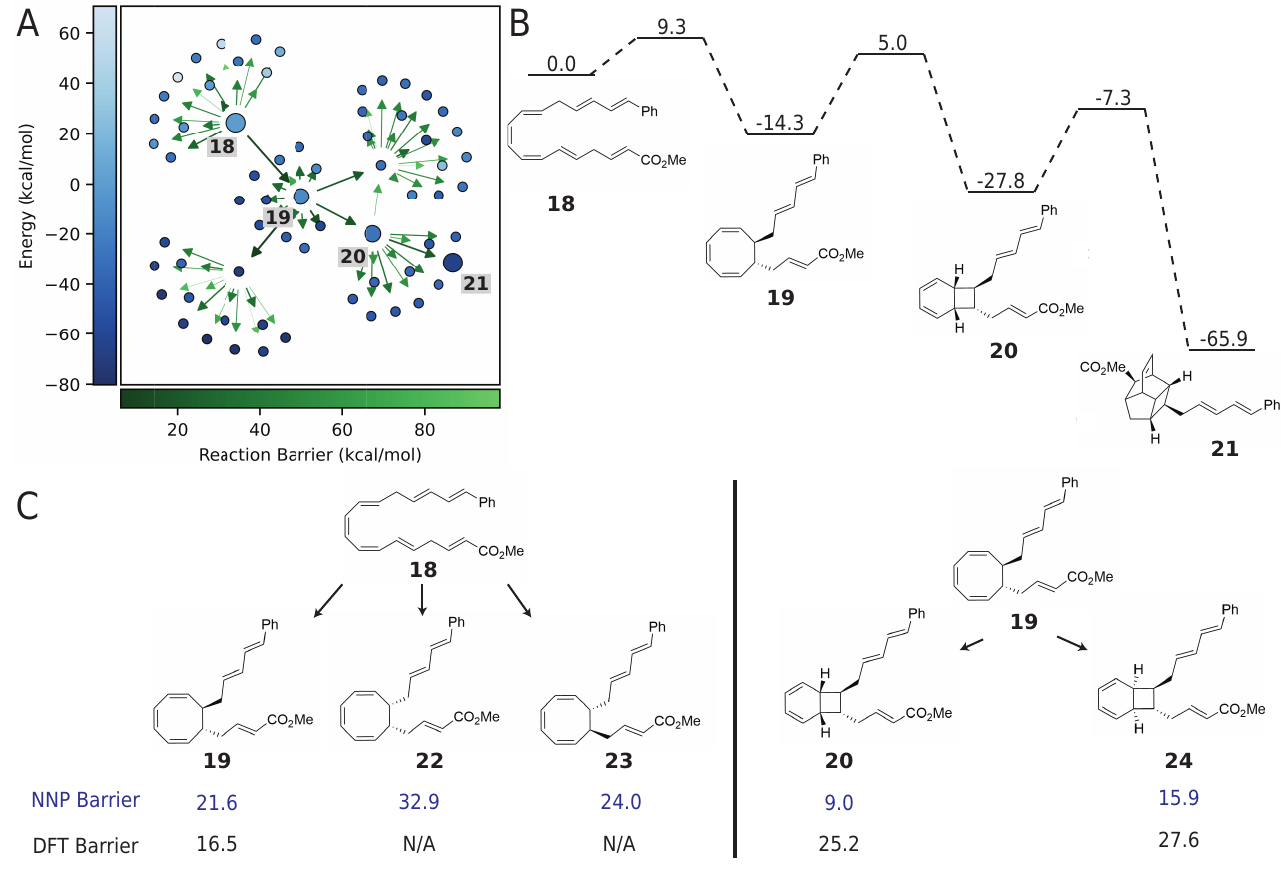}
    \caption{Analysis of the synthesis of endiandric acid C. \textbf{A.} \crncaption{17} \textbf{B.} A kinetically favorable pathway identified by REVAMP corresponds to the synthesis performed by \citeauthor{nicolaou_endiandric_1982} involving multiple stereoselective cyclizations with many simultaneous bond breaking/forming events. The energies of each state are reported in kcal/mol normalized to the reactants. \textbf{C. } The AIMNet2-rxn and DFT calculated barriers for the favored electrocyclizations along with their less favored stereoisomers. A value of N/A means a transition state wasn't localized for the given reaction. The barriers are reported in kcal/mol normalized to the immediate precursor.}
    \label{fig:endiandric}    
\end{figure}

From the reactant \textbf{18}, the first step that is found is an 8$\pi$ contrarotary cyclization to form an 8 membered ring (\textbf{19}). This is followed by a 6$\pi$ disrotary cyclization. Finally, a Diels-Alder reaction is used to form the bridged ring that is in the product (\textbf{21}). With the stereoisomers enumerated from REVAMP, we are able to successfully determine the correct stereoisomer for each of these steps using the calculated barriers (Figure \ref{fig:endiandric}C). The accuracy of AIMNet2-rxn compared to the reference DFT can vary depending on the chemical space and reaction mechanism coverage. We therefore use DFT as the final evaluation during each single-step exploration to provide a consistent physics-based evaluation; importantly, the ranking of outcomes remains consistent, empirically. Another justification for the inclusion of DFT as a final evaluation are the N/A barrier values in Figure 5. These indicate that the DFT transition state optimization did not converge to a valid transition state which means that either the AIMNet2-rxn transition state conformer was too far away from the corresponding DFT transition state conformer or that the corresponding DFT transition state conformer doesn't exist. In the former case, increasing the number of transition states sampled by AIMNet2-rxn or incorporating global optimization techniques (like Monte Carlo multiple minimum sampling \cite{chang_internal-coordinate_1989}) would help increase the reliability of DFT transition state convergence at an increased computational expense. In the latter case, we assume that the DFT PES is a better representation of the ground truth PES thus allowing us to label these transition states and their corresponding elementary steps as kinetically irrelevant.

\newpage
\begin{table}[htbp]
\caption{Summary of REVAMP's exploration behavior for each of the six case studies. Columns indicate the number of atoms in the starting reactants (N); the minimum required ``n'' necessary in the ``bnfn'' formalism to recover the experimentally-reported pathways (bnfn); the numbers of intermediates considered at each stage of filtering, starting from the initial enumeration (Total), after the rapid 2D assesment of stability (Stability), after the rapid 2D assessment of thermodynamic feasibility ($\Delta E_{rxn}^{MPNN}$), after the AIMNet2-rxn estimate of reaction energy ($\Delta E_{rxn}^{NNP}$), then the AIMNet2-rxn estimate of kinetic feasibility ($\Delta E^{\dagger, NNP}$) and finally DFT ($\Delta E^{\dagger, DFT}$) (the number of converged DFT transition states); the total wall time to run the full pipeline excluding confirmatory DFT calculations (Pre-DFT), and the total wall time to run all DFT calculations (DFT). Computational times are based on parallelized calculations of 7 nodes with 48 cores each}
\label{tab:summary}
\resizebox{\linewidth}{!}{
\begin{tabular}{@{}lrrrrrrrrrr@{}}
\toprule
\multicolumn{3}{l|}{}                              & \multicolumn{6}{c|}{Intermediates after filtering}                                                                                                          & \multicolumn{2}{c}{Wall time (h)} \\ \midrule
Case Study        & N  & \multicolumn{1}{l|}{bnfn} & Total  & Stability & $\Delta E_{rxn}^{MPNN}$ & $\Delta E_{rxn}^{NNP}$ & \multicolumn{1}{l}{$\Delta E^{\dagger,NNP}$} & \multicolumn{1}{l|}{$\Delta E^{\dagger,DFT}$} & Pre-DFT          & DFT            \\ \midrule
Benchmark - 1     & 20 & 3                         & 14120  & 8714      & 3515                 & 750                  & 545                                        & 97                                           & 5.1              & 21.3           \\
Benchmark - 2     & 24 & 4                         & 4030   & 1995      & 1880                 & 1206                 & 198                                        & 60                                           & 0.7              & 14.6           \\
DA - 1            & 33 & 3                         & 49548  & 38006     & 14357                & 2217                 & 531                                        & 22                                           & 6.0              & 36.9           \\
DA - 2            & 34 & 3                         & 80290  & 53683     & 5812                 & 514                  & 173                                        & 20                                           & 2.1              & 57.7           \\
Salvinorin A      & 36 & 3                         & 23582  & 15548     & 12571                & 5281                 & 690                                        & 28                                           & 13.1             & 23.7           \\
Endiandric acid C & 52 & 4                         & 911946 & 775230    & 521271               & 85068                & 402                                        & 146                                           & 62.8             & 298.7          \\ \bottomrule
\end{tabular}
}
\end{table}

\section{Conclusion}

In this article, we have demonstrated the productive combination of graph-based enumeration and neural network potentials for the computational evaluation of complex cyclization reactions with relevance to natural product synthesis planning. We implement our workflow in an open-source tool, REVAMP, which by default performs a ``b4f4'' and stereoenumeration scheme to generate a broad range of potential outcomes. To cope with the large number of possible elementary steps, we employ inexpensive machine learning models for intermediate filtering and perform subsequent thermodynamic and kinetic assessment with the reactive NNP AIMNet2-rxn. Many elements of this workflow take direct inspiration from YARP. We illustrate the successful construction of complex cyclization pathways  with various case studies: a stereoselectivity study of an intramolecular Diels-Alder reaction and two examples from natural product syntheses representing various levels of elementary step complexity. 

The summary statistics presented in Table \ref{tab:summary} demonstrate REVAMP's ability to explore these mechanisms with manageable computational cost. One clear trend that emerges is that the number of enumerated intermediates is very sensitive to the number of atoms which compounds the existing challenges associated with modeling larger reacting systems. REVAMP partially mitigates this combinatorial explosion with its progressive filtering stages which help reduce the number of DFT transition state optimizations by at least an order of magnitude in all case studies. Despite the efficacy of these filters, the wall time associated with performing the DFT calculations is still quite high in cases with larger molecules, highlighting the challenges associated with modeling systems of this size and complexity.

The rise of NNPs capable of providing DFT-quality energies for equilibrium and transition state geometries will continue to assist mechanistic modeling. At present, the strength of REVAMP is in the analysis of neutral, closed-shell reactions, particularly pericyclic reactions like electrocyclizations and cycloadditions. Currently, REVAMP is incapable of handling elements outside of C, H, O, and N, charged species, and open-shelled chemistry. This represents a significant limitation of its utility beyond analyzing the particular reaction types in this study. However, due to the modular nature of REVAMP, it will be straightforward to exchange the current NNP for those that will be released in coming years. In particular, we expect models trained on datasets like OMol25 \cite{levine_open_2025} as well as recently published AIMNet2 models, AIMNet2-Pd \cite{anstine_transferable_2025} and AIMNet2-NSE \cite{kalita_aimnet2-nse_2025}, to help drive the modeling of an increased reaction scope. A key consideration to ensure successful integration of these models will be ensuring these models generalize well across chemical reactions by performing activation barrier benchmarks as done in this manuscript as well as in the AIMNet2-rxn manuscript \cite{anstine_aimnet2-rxn_2025}.

\begin{acknowledgement}

This material is based upon work supported by the National Science Foundation under Grant No. CHE-2202693. 

\end{acknowledgement}

\begin{suppinfo}

Additional implementation details and more detailed results of the pathways identified in each case study.

\end{suppinfo}

\begin{datastatement}

The code and data that support the findings of this study are openly available at 
\url{https://github.com/ncasetti7/REVAMP/tree/main}. The neural network potential used in this study will be available when its corresponding paper \cite{anstine_aimnet2-rxn_2025} is published.

\end{datastatement}

\bibliography{references}

\end{document}


\newcommand{\crncaption}[1]{The chemical reaction network returned by REVAMP starting from \textbf{#1} where arrow color represents barrier height and node color represents intermediate energy relative to the reactant. Larger nodes correspond to species along the reaction pathway}

\newcommand{\doublecrncaption}[2]{The chemical reaction network returned by REVAMP starting from \textbf{#1} and \textbf{#2} respectively where arrow color represents barrier height and node color represents intermediate energy relative to the reactant. Larger nodes correspond to species along the reaction pathway}

\newcommand{\fullnetworkcaption}[1]{The chemical reaction network returned by REVAMP for case study #1 where arrow color represents barrier height. Solid arrows represent the experimentally observed reaction pathway.}

\newpage

\section{Graph-based enumeration demonstration}
\label{bnfn}

\begin{figure}[h]
    \centering
    \includegraphics[width=\textwidth]{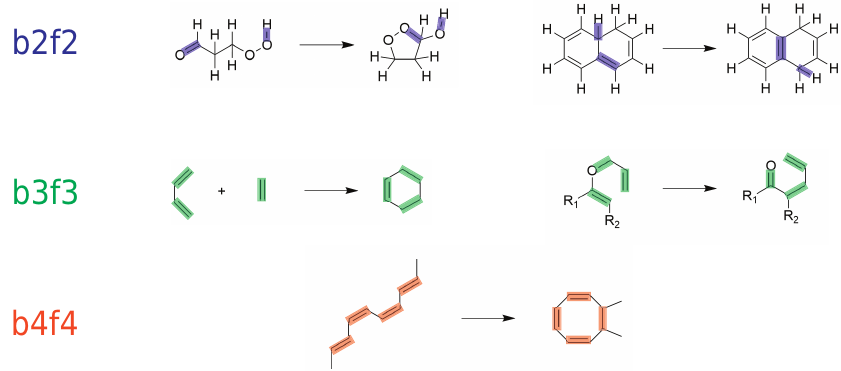}
    \caption{Examples of reactions enumerated by graph-based enumeration ordered by the number of bonds broken and formed.}
    \label{fig:bnfn}
\end{figure}

\newpage

\section{Intermediate pool size for various enumeration schemes}\label{enumeration}

\begin{figure}
    \centering
    \includegraphics[width=\textwidth]{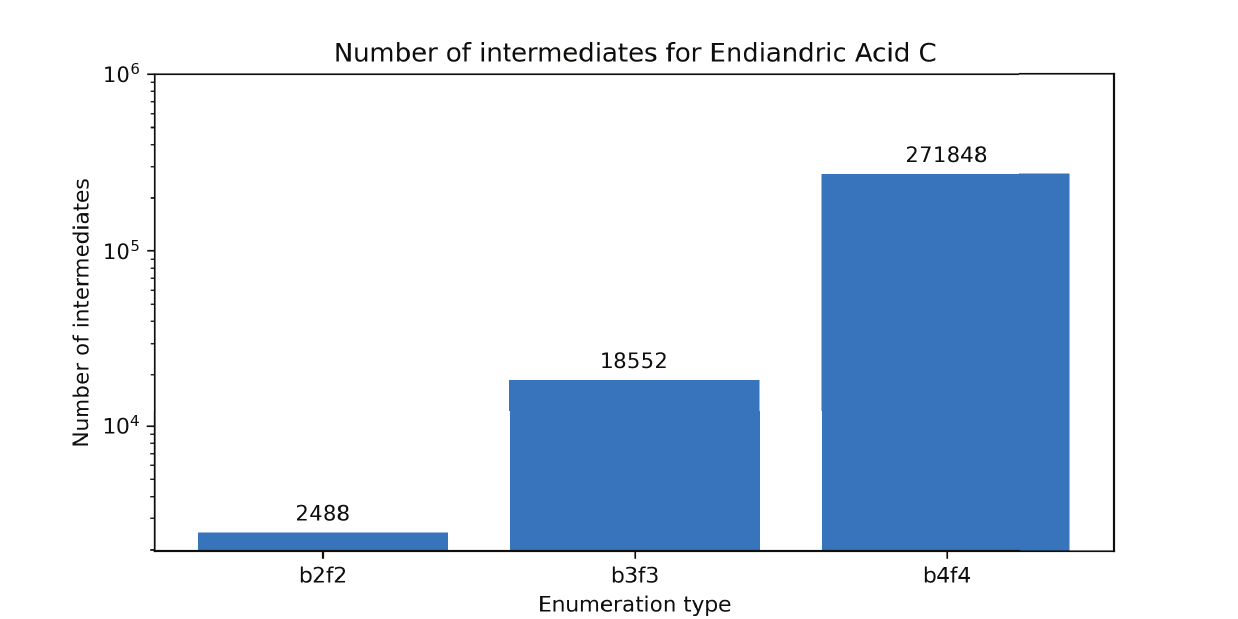}
    \caption{Growth of intermediate pool for larger enumeration schemes for Endiandric Acid C}
    \label{enumeration_size}
\end{figure}

\newpage

\section{Conformer generation and optimization cost breakdown}\label{conformer}

Benchmarking studies performed using the ETKDG method \cite{wang_improving_2020} followed by UFF \cite{rappe_uff_1992} optimization have found that conformer generation operates between 0.1 - 1 CPU seconds\cite{seidel_high-quality_2023}. Optimizations with the NNP operate around 0.01 CPU seconds per optimization step with the maximum number of cycles set to 300. The combination of these two operations results in an approximate run time on the order of 1 CPU seconds.

\newpage

\section{Additional details of methods}
\label{detail}

In order to make graph-based enumeration tractable, conditional enumeration is used for b3f3 and b4f4 elementary steps. Conditional enumeration is a technique pioneered by YARP where a b3f3 or b4f4 elementary step is only considered valid if it includes a double bond which is an assumption that has been shown to retain kinetically relevant intermediates\cite{zhao_algorithmic_2022}. The data for the training of the stability classification and reaction energy regression MPNNs was generated by performing graph-based enumeration on 2100 reactants randomly (constrained to neutral molecules with only C, H, O, and N) sampled from reactions from the United States Patent and Trademark Office (USPTO) \cite{lowe_extraction_2012}, randomly selecting 1000 of these intermediates, and using the NNP to evaluate their stability and reaction energy as described in the main text. The models were trained using Chemprop 1.7.0 \cite{yang_analyzing_2019, heid_machine_2022} with the default settings. Data was split randomly 80-10-10 for train-validation-test for the model used in this manuscript. When assessing kinetic feasibility, four reactant-product conformer pairs are selected and evaluated from the conformer pairs generated by the method described in the main text of the paper. All of the transition states corresponding to a reaction (as determined by the IRC) are saved, but for a given reaction, only different transition states (as defined by having greater than 0.5 angstrom RMSD) are considered for processing with DFT. For duplicate transition states, the lower energy transition state is selected. Of these different transition states, all transition states below a predefined threshold (defaulting to 60 kcal/mol) are passed to DFT calculations which are performed as described in the main text.

\newpage

\section{Performance of message passing neural networks}
\label{mpnn}

The figure below outlines the performance of the two MPNNs on a randomly selected test set from the dataset described in the previous section. The binary classification model does a solid job of filtering out unstable species while minimizing the number of false negatives (a stable species labelled as unstable). The regression model struggles with numerical accuracy but is generally capable of maintaining rank order of candidate intermediates.

\begin{figure}
    \centering
    \includegraphics[width=\textwidth]{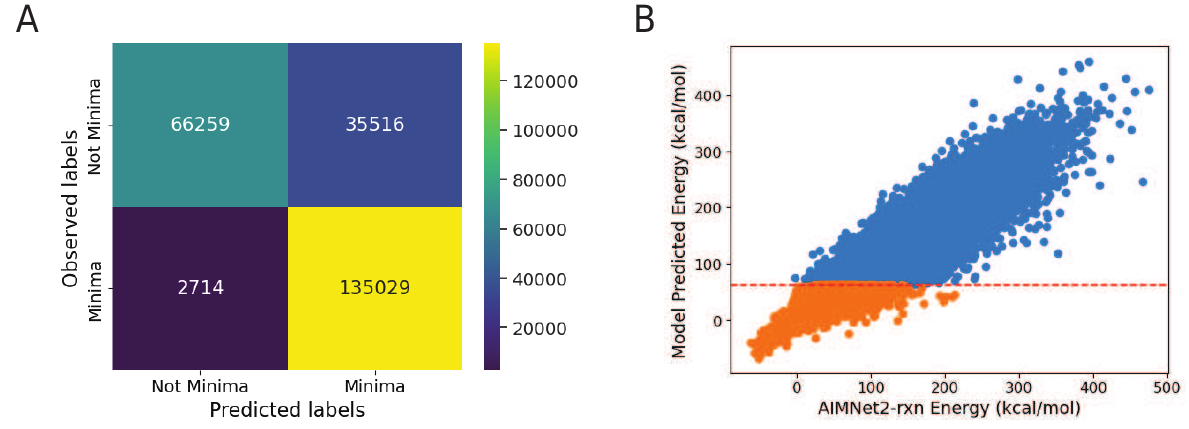}
    \caption{A) Confusion matrix for the binary classification model on a random test set where "not a minima" represents species that don't conserve their bonding matrix during optimization and "minima" represents the opposite. This model achieves solid accuracy and a low false negative rate. B) Parity plot for the regression model on a random test set. Although accuracy is somewhat poor, the model maintains decent rank order of candidates. The red line represents an example threshold where points below would be passed to the next step and points above would be discarded.}
    \label{fig:mpnn}
\end{figure}

\newpage

\section{Default mechanism search energetic thresholds}
\label{threshold}

The thresholds in the table below are the thresholds used to perform the mechanism searches in this manuscript. These thresholds are relatively permissive as we wanted to explore a wide range of transformations in this manuscript. Tightening these thresholds to values that better reflect kinetic relevance (i.e. reaction energy thresholds of $\sim$10 kcal/mol and a barrier threshold of $\sim$30 kcal/mol) would significantly reduce computational expense but also increase risk of filtering out kinetically relevant transformations.

\begin{table}[]
\caption{Threshold values used to filter intermediates for the searches performed in this manuscript. Whichever threshold was more restrictive between the top-k threshold and the AIMNet2-rxn barrier threshold is what was used to determine the number of TS structures passed for DFT refinement.}
\label{tab:threshold}
\begin{tabular}{@{}llll@{}}
\toprule
Threshold            & Description                           & Default Value & Unit     \\ \midrule
$E_{MPNN}$           & MPNN reaction energy threshold        & 60            & kcal/mol \\
$E_{NNP}$            & AIMNet2-rxn reaction energy threshold & 30            & kcal/mol \\
$E^{\ddagger}_{NNP}$ & AIMNet2-rxn barrier threshold         & 60            & kcal/mol \\
k                    & Maximum number of TSs passed to DFT   & 100           &          \\ \bottomrule
\end{tabular}
\end{table}

\newpage

\section{Speed benchmark of NNPs and GFN2-xTB}
\label{speed}

NNPs trained on the recently released OMol25 dataset \cite{levine_open_2025}, esen-small and esen-medium, provide significantly slower inference than AIMNet2-rxn and GFN2-xTB making them less suited for the REVAMP workflow as reduced computational expense enables more thorough conformational sampling.

\begin{figure}[h]
    \centering
    \includegraphics[width=0.5\textwidth]{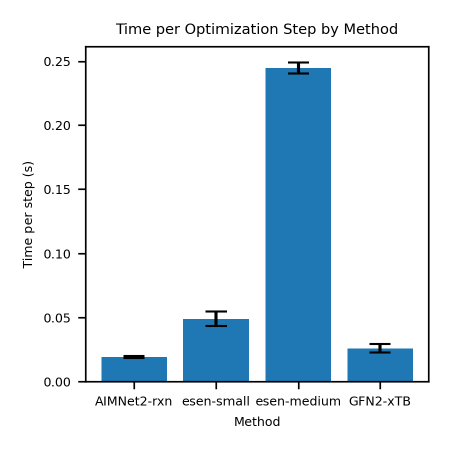}
    \caption{Time per optimization step for optimization of molecule \textbf{1} from a random conformer generated as described in the main text. Error bars represent the standard deviation of average optimization step time over 5 repeat optimizations.}
    \label{fig:time-per-cycle}
\end{figure}

\newpage

\section{Benchmark mechanism search details}

\begin{figure}[h!p]
    \centering
    \includegraphics[width=\textwidth]{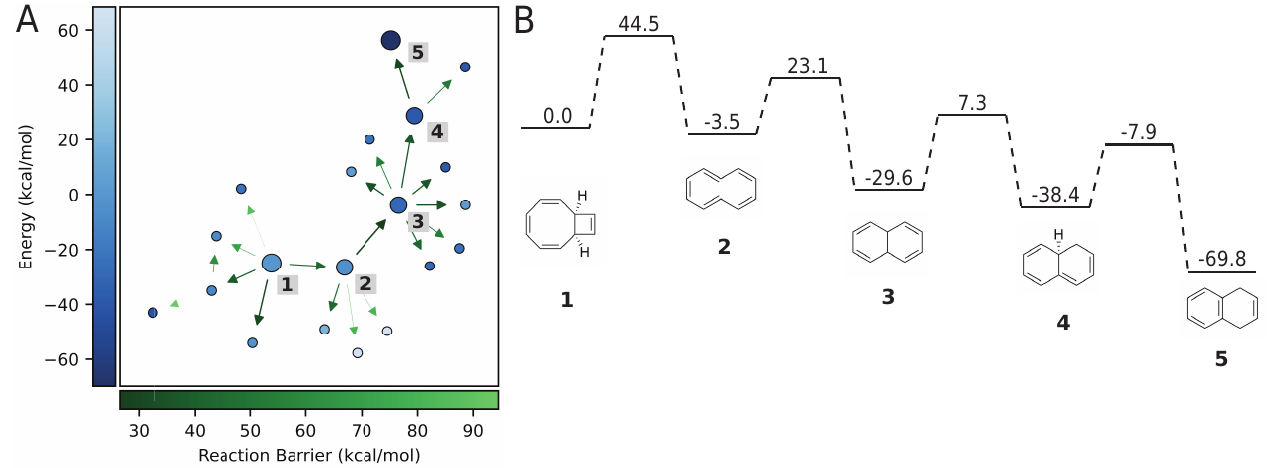}
    \caption{\textbf{A.} \crncaption{6}. \textbf{B.} A kinetically favorable pathway identified by REVAMP involving multiple sequential cyclizations and the stereoselective formation of a fused ring system. }
    \label{fig:electrocyclic}
\end{figure}

\begin{figure}[h!p]
    \centering
    \includegraphics[width=\textwidth]{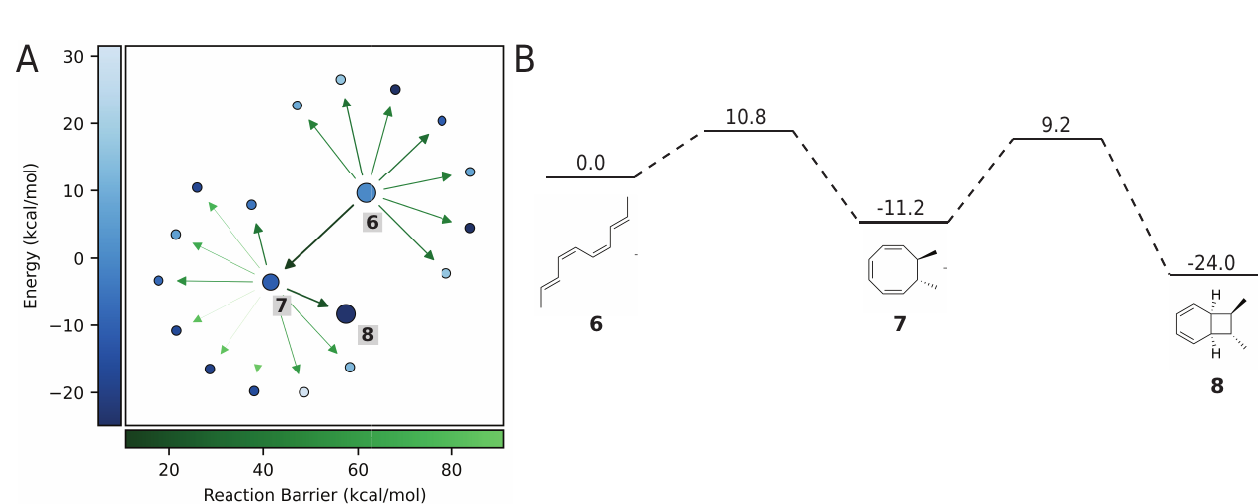}
    \caption{\textbf{A.} \crncaption{11}. \textbf{B.}  A kinetically favorable route identified by REVAMP with steps involving 4 and 3 simultaneous bond breaking/forming events, which could not have been found by a graph-based method limited to b2f2 reactions.}
    \label{fig:enter-8pi_electrocyc}
\end{figure}

\newpage

\section{Full reaction networks for case studies}

\begin{figure}[hp]
    \centering
    \includegraphics[width=\textwidth]{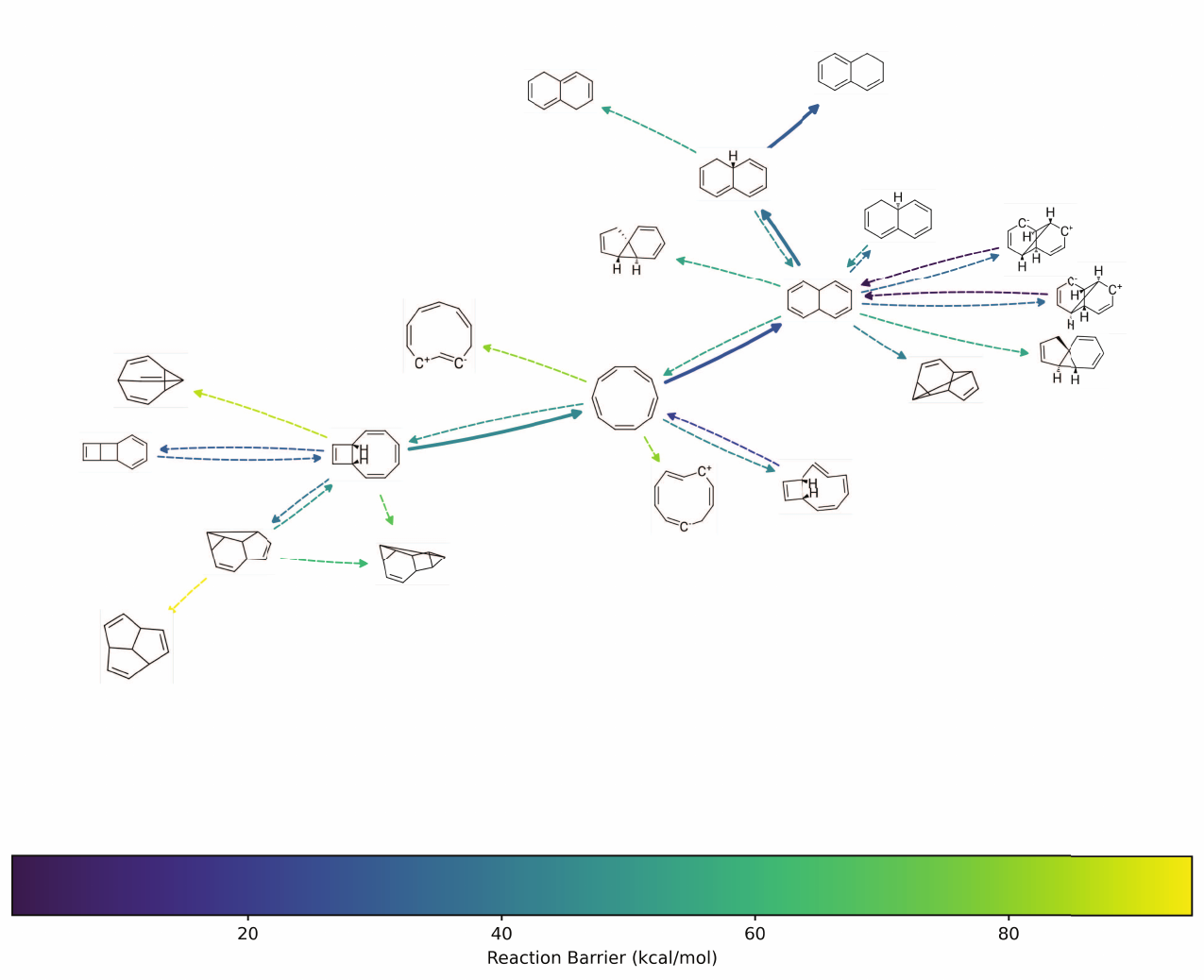}
    \caption{\fullnetworkcaption{Benchmark 1}}
    \label{fig:full_electrocyclic}
\end{figure}

\begin{figure}[hp]
    \centering
    \includegraphics[width=\textwidth]{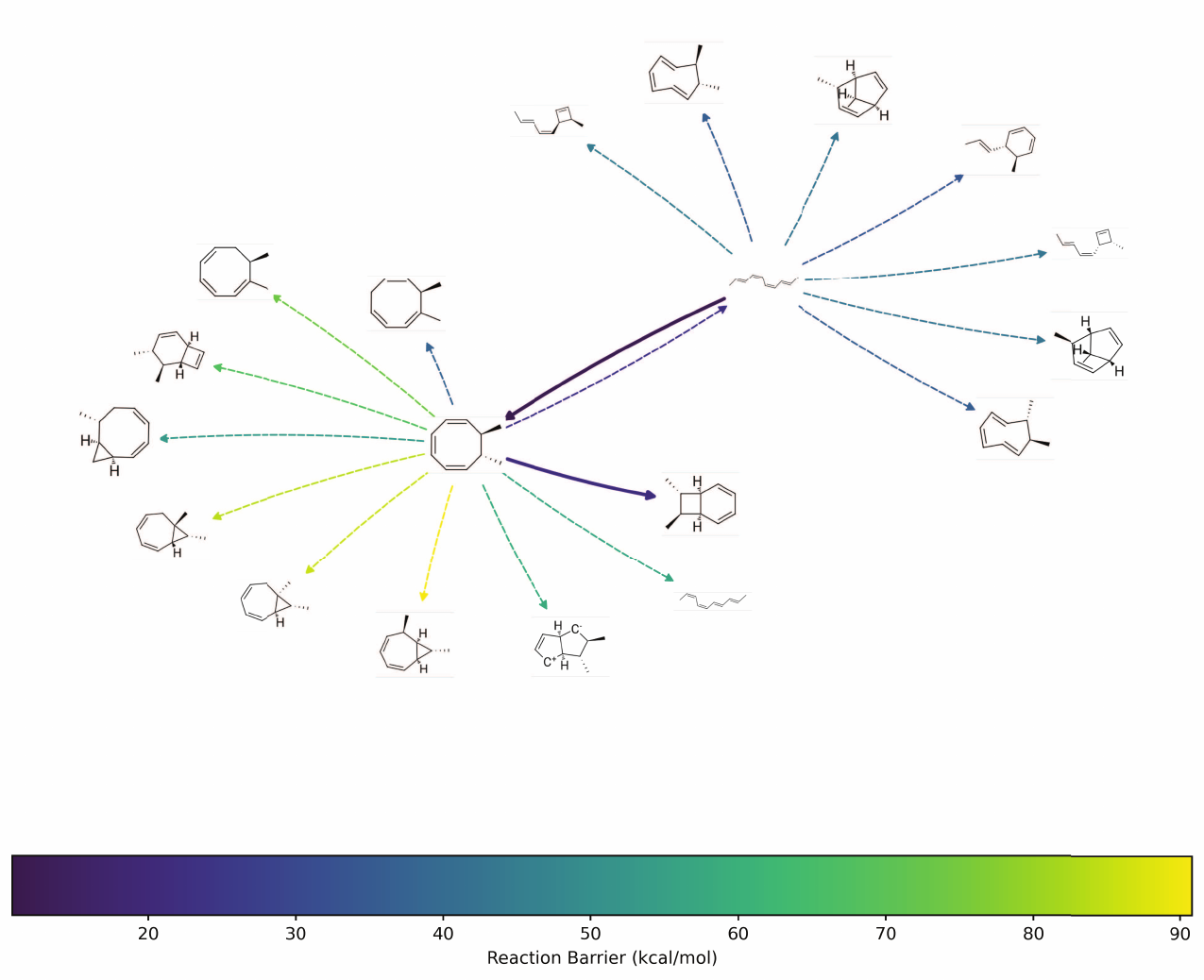}
    \caption{\fullnetworkcaption{Benchmark 2}}
    \label{fig:full_8pi}
\end{figure}

\begin{figure}[hp]
    \centering
    \includegraphics[width=\textwidth]{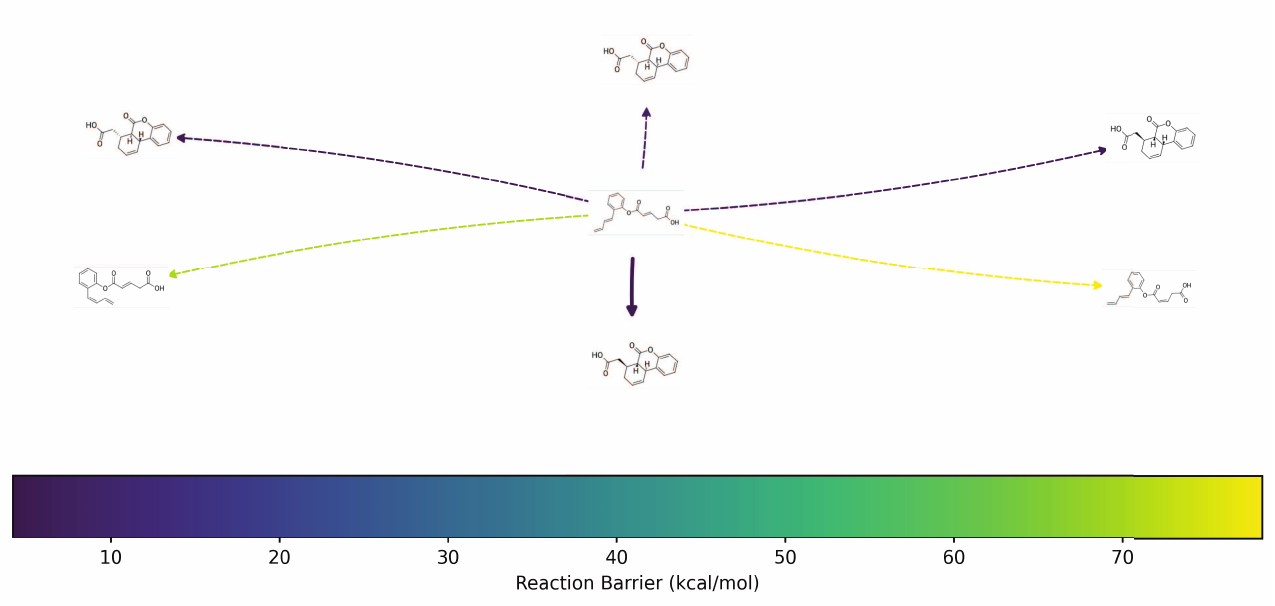}
    \caption{\fullnetworkcaption{DA 1}}
    \label{fig:full_da1}
\end{figure}

\begin{figure}[hp]
    \centering
    \includegraphics[width=\textwidth]{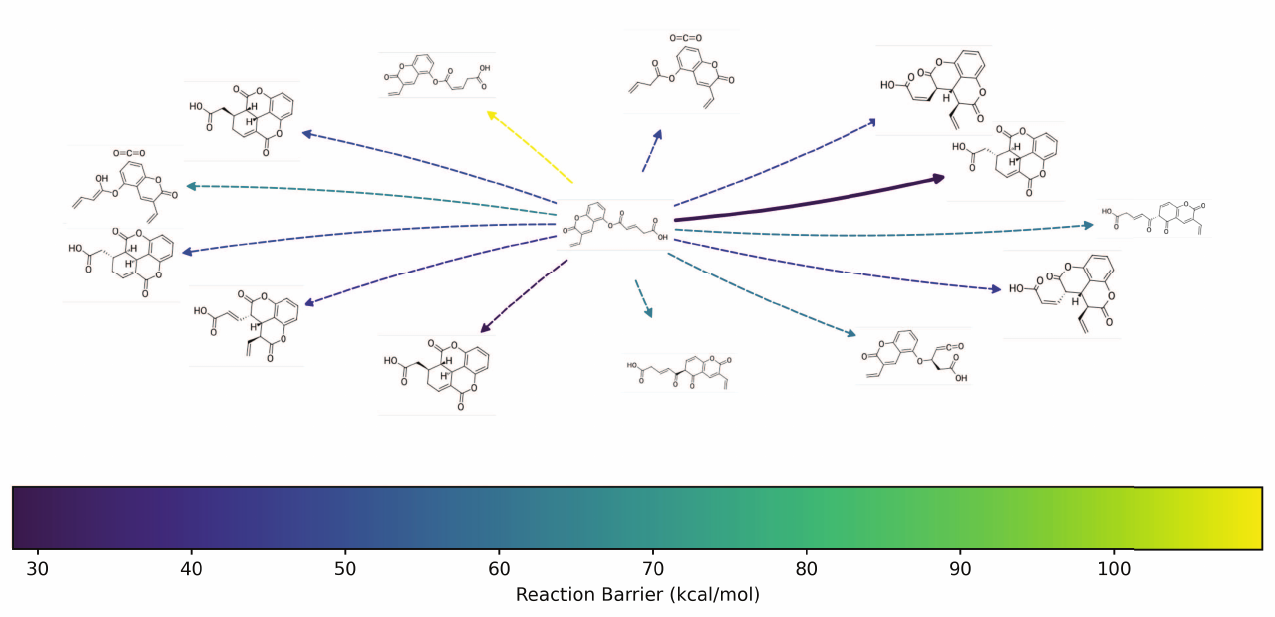}
    \caption{\fullnetworkcaption{DA 2}}
    \label{fig:full_da2}
\end{figure}

\newpage

\begin{figure}[hp]
    \centering
    \includegraphics[width=\textwidth]{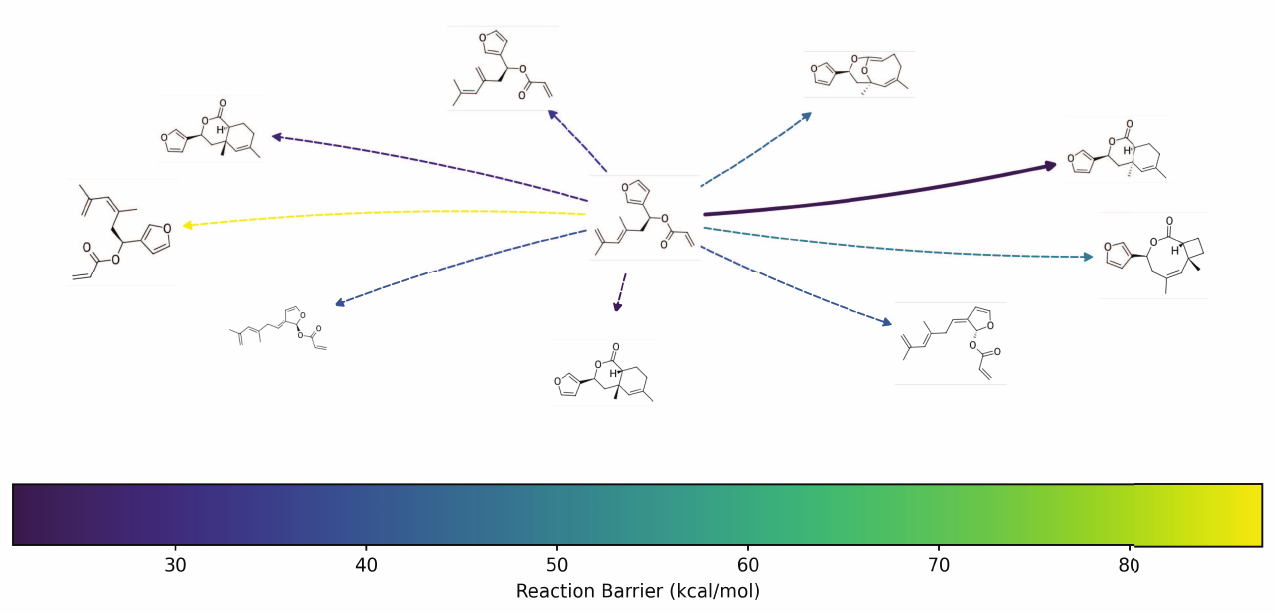}
    \caption{\fullnetworkcaption{Salvinorin A}}
    \label{fig:full_salvinorin}
\end{figure}

\newpage

\bibliography{references}